\documentclass{article}

\usepackage{multirow}

 \usepackage[preprint]{neurips_2026}

\usepackage{amsmath,amssymb}
\numberwithin{equation}{section}
\usepackage{fullpage}
\usepackage{graphicx}
\usepackage{float}
\usepackage{url}
\usepackage{booktabs}
\usepackage{xcolor}
\usepackage{hyperref}
\usepackage{cleveref}
\usepackage{algorithm}
\usepackage{algorithmic}

\begin{document}

\title{Spectral Distillation: From Nonlinear Dynamics to Linear State-Space Models}

\author{
  Liane Galanti\thanks{
    Equal contribution; author order chosen alphabetically.
    Corresponding emails:
    \texttt{galanti@princeton.edu},
    \texttt{devan.shah@princeton.edu}.
  } \\
  Princeton University
  \And
  Devan Shah\footnotemark[1] \\
  Princeton University
  \And
  Shlomo Fortgang \\
  Princeton University
  \And
  Elad Hazan \\
  Princeton University
}

\maketitle

\begin{abstract}
Can nonlinear dynamical systems be learned through a compact linear state-space representation, without directly solving a non-convex system-identification problem? We give a provable pipeline for doing so. Starting from observations of an unknown nonlinear dynamical system, we first learn an implicit spectral predictor using Observation Spectral Filtering (OSF), a convex method that competes with the best linear observer for the system. We then apply spectral-to-LDS distillation to convert this predictor into an explicit recurrent linear dynamical system. Our main theorem shows that the average prediction error of the distilled LDS decomposes into an exponentially-small distillation term and the OSF learning term governed by the Luenberger complexity of the best observer. The guarantee is dimension-free: it depends on observer complexity rather than on the latent dimension needed to represent the nonlinear system. To our knowledge, this yields the first end-to-end provable method for extracting a best-in-hindsight LDS representation of nonlinear dynamics through convex learning followed by provable distillation. Experiments on linear LDS benchmarks and MuJoCo behavior cloning show that the train-then-distill pipeline produces compact LDS predictors that match or outperform directly trained baselines.

\end{abstract}

\section{Introduction}

Modern sequence modeling and control hinge on the ability to represent and learn dynamical systems. Linear dynamical systems (LDS) occupy a central role in this landscape: they admit compact state-space representations, efficient simulation, and a rich spectral theory that enables both analysis and algorithm design. However, most real-world systems are inherently nonlinear, and learning either a nonlinear state-space model or even its best linear approximation remains a fundamentally non-convex and fragile problem.

This paper asks a basic question:
\emph{can we extract a simple linear dynamical system that faithfully represents a nonlinear one, without ever solving a non-convex system identification problem?}
 
Our answer is affirmative. We present a provable nonlinear-to-LDS distillation pipeline that converts observations from an unknown nonlinear system into a compact linear dynamical system, using only convex optimization followed by a structured distillation step.
 
The key idea is to separate {learning} from {representation}. Rather than attempting to fit a state-space model directly, we first learn an implicit predictor using Observation Spectral Filtering (OSF), a convex method that operates in a fixed temporal feature space. OSF does not attempt to recover latent states; instead, it competes with the best linear observer that could explain the data, achieving regret bounds that depend only on an intrinsic measure of observability complexity.
 
We then convert this predictor into an explicit state-space model. Leveraging recent results on spectral representations of dynamical systems, we show how a learned spectral predictor can be distilled into a finite-dimensional LDS whose impulse responses approximate those of the predictor up to exponentially small error. The result is a standard recurrent model with memory and per-step cost depending only on the observation and output dimension, suitable for long-horizon simulation and control.

\subsection{Our Contributions}

This paper makes the following contributions.

\begin{itemize}
    \item \textbf{A nonlinear-to-LDS distillation pipeline.}
    We give a two-stage procedure that first learns a nonlinear dynamical system through the convex OSF feature class, and then distills the resulting spectral predictor into an explicit recurrent LDS.

    \item \textbf{A dimension-free prediction guarantee.}
    We prove that with $k=\Theta(\log^2 T)$ spectral filters and $h \geq k$ SpectraLDS poles, the distilled LDS satisfies
           \[
        \frac{1}{T}\sum_{t=1}^T
        \|y_t-\hat y_t^{\mathrm{LDS}}\|^2
        \le
        \widetilde O\!\left(
        \frac{Q_\star^2\log(Q_\star)\log^7T}{\sqrt T}
        +
        \lambda_{\max}(F)^2
        \exp\!\left(-\frac{2k}{\log T}\right)
        \right).
        \]
    Thus the error decomposes into an exponentially small distillation term and the OSF learning term governed by the Luenberger complexity $Q_\star$.

    \item \textbf{An explicit compact representation.}
    Unlike OSF, which is an implicit convolutional predictor, the output of our method is a standard LDS with $O(h \cdot (d_{in}+d_{out}))$ memory and $O(h \cdot (d_{in} + d_{out}))$ per-step simulation cost. The guarantee depends on observer complexity rather than on the potentially enormous dimension of a nonlinear lift or discretization.

    \item \textbf{Empirical validation of train-then-distill.}
    We test the pipeline on linear LDS benchmarks and MuJoCo behavior cloning. Across these settings, the distilled LDS preserves the behavior of the spectral predictor and often outperforms directly trained LDS baselines.
\end{itemize}

\noindent
Together, these results give a principled route from nonlinear observations to compact linear state-space models: learn in a convex spectral space, then distill into an interpretable recurrent LDS.

\subsection{Related Work}
\label{subsec:related}

\paragraph{Linear dynamical system identification.}
Classical system identification \cite{ljung1999system} fits state-space parameters $(A,B,C)$ directly, typically through non-convex optimization. Recent work has obtained non-asymptotic guarantees for LDS estimation under Gaussian noise \cite{hardt2018gradient, simchowitz2018learning, oymak2019nonasymptotic, sarkar2019near, tsiamis2019finite}, with sample complexity scaling polynomially in the state dimension and condition number of the system. These results crucially assume the data are generated by a linear system; extending them to nonlinear settings has remained open.

\paragraph{Spectral filtering.}
Spectral filtering \cite{hazan2017spectral, hazan2018spectral} sidesteps non-convex identification by learning in a fixed convolutional feature space derived from the Hankel matrix. The OSF extension \cite{dogariu2025universal} combines this with a Luenberger-observer analysis to obtain regret guarantees for nonlinear systems through improper learning. SpectraLDS \cite{shah2025spectralds} converts spectral predictors into recurrent LDS form. Our work composes these ingredients into an end-to-end provable pipeline with a unified guarantee, and we describe these techniques more in the next section.

\paragraph{Koopman operator methods.}
Koopman theory \cite{mezic2005spectral} represents nonlinear dynamics as a linear operator on an infinite-dimensional observable space, and finite-dimensional approximations underlie methods such as EDMD \cite{williams2015data} and deep Koopman embeddings \cite{lusch2018deep, kostic2022learning}. These approaches typically require choosing a dictionary of observables and offer mostly asymptotic or empirical guarantees. Our pipeline can be viewed in a similar spirit, but the spectral-filter dictionary is universal and the guarantees are non-asymptotic.

\paragraph{Modern state-space sequence models.}
Structured state-space models such as S4 \cite{gu2022efficiently}, S5 \cite{smith2023simplified}, the LRU \cite{orvieto2023resurrecting}, and Mamba \cite{gu2023mamba} have shown that linear recurrences with carefully chosen parameterizations can match or exceed Transformer performance on long-sequence tasks. These works focus on architectural and empirical advances rather than provable system-identification guarantees. Our distilled LDS produces a compact recurrence in the same family, but with a provable approximation guarantee for nonlinear data-generating processes.

\section{Background}
\label{sec:background}

We recall the two ingredients used by our pipeline. Observation Spectral Filtering (OSF) \cite{dogariu2025universal} learns a convex spectral predictor \cite{hazan2017spectral} from past inputs and observations. Spectral-to-LDS distillation \cite{shah2025spectralds} then realizes the learned spectral predictor as an explicit recurrent state-space model.

\subsection{Linear systems as convolutional predictors}

A controlled linear dynamical system maps inputs $u_t\in\mathbb{R}^{d_u}$ to outputs $y_t\in\mathbb{R}^{d_y}$ through a hidden state $x_t\in\mathbb{R}^{d_x}$:
\begin{equation}
    x_{t+1} = A x_t + B u_t,
    \qquad
    y_t = C x_t + D u_t .
    \label{eq:lds-background}
\end{equation}
The direct term $D u_t$ can be replaced by an expanded hidden state, and so we often consider $D = 0$. Taking $x_0=0$ and suppressing the direct term, the output expands as
\begin{equation}
    y_t
    =
    \sum_{\ell=0}^{t-1} C A^\ell B\,u_{t-\ell}.
    \label{eq:lds-convolution-background}
\end{equation}
Thus an LDS is equivalently described by its impulse response
$\{C A^\ell B\}_{\ell\ge 0}$.

This representation exposes the main difficulty in directly fitting LDSs. Long-memory systems require accurately learning high powers of $A$, and the map from $A$ to the sequence $\{A^\ell\}_{\ell\ge 0}$ is non-convex and poorly conditioned when $\rho(A)$ is close to one. Spectral filtering \cite{hazan2017spectral} avoids this difficulty by learning in a fixed convolutional feature space instead of identifying the latent transition matrix.

\subsection{Spectral filtering}

For $\alpha\in[0,1]$, define the length-$L$ geometric response
\[
    \mu_L(\alpha)
    =
    (1,\alpha,\alpha^2,\ldots,\alpha^{L-1}) .
\]
Consider the Hankel spectral basis obtained from
\begin{equation}
    Z_L
    =
    \int_0^1
    \mu_L(\alpha)\mu_L(\alpha)^\top\,d\alpha .
    \label{eq:hankel-background}
\end{equation}
Let $\phi^+_1,\ldots,\phi^+_k$ be the top $k$ eigenvectors of $Z_L$, with corresponding eigenvalues
$\sigma_1,\ldots,\sigma_k$. The Spectral Filtering  algorithm \cite{hazan2017spectral} is rooted in the observation that the span of these filters uniformly approximates length-$L$ geometric responses, with approximation error decaying exponentially in $k/\log L$.

For a signal stream $s_t$, define the spectral features
\begin{equation}
    U_{t,j}^{(s)}
    =
    \left\langle \phi_j^+, s_{t-1:t-L}\right\rangle
    =
    \sum_{\ell=1}^{L}\phi^+_j(\ell)\,s_{t-\ell},
    \qquad j=1,\ldots,k .
    \label{eq:spectral-features-background}
\end{equation}
A spectral predictor over this stream has the form
\begin{equation}
    \hat y_t^{\mathrm{SF}}
    =
    \sum_{r=1}^{m} R_r s_{t-r}
    +
    \sum_{j=1}^{k} M_j U_{t,j}^{(s)} .
    \label{eq:sf-background}
\end{equation}
The temporal filters $\phi^+_j$ are fixed in advance, while the readout matrices $R_r$ and $M_j$ are learned. Squared loss is therefore convex in the trainable parameters, even though the resulting predictor can represent long-memory impulse responses for an LDS with PSD matrix $A$. To extend this result to a symmetric LDS \cite{hazan2018spectral}, where the transition matrix $A$ is symmetric, we can expand our filter set to include the negative spectral filters $\phi_i^{-} = [(\phi^+_i)_t \cdot (-1)^t]_{t=0}^{L-1}$. Thus, when we refer to $k$ spectral filters $\phi_i$, we refer to $k/2$ positive filters $\phi^+_i$ and $k/2$ negative filters $\phi_i^{-}$. For ease of explanation, we redefine the spectral features and predictor to use the combined filter set $\phi_1, \dots \phi_k$.

\subsection{Observation Spectral Filtering}

OSF \cite{dogariu2025universal} extends spectral filtering to the input-output prediction setting. At time $t$, the learner predicts $y_t$ from both the control history and the observation history. The form of the predictor is motivated by the Luenberger observer for a controlled linear system:
\[
    \hat x_{t+1}
    =
    A\hat x_t + B u_t + L(y_t-C\hat x_t);\quad \hat{y}_t = C\hat{x}_t
\]
The observer error evolves according to the closed-loop matrix
\[
    \widetilde A = A-LC .
\]
Expanding the observer prediction expresses $y_t$ as a combination of recent inputs, recent observations, and long convolutions over both streams, with convolution kernels generated by powers of $\widetilde A$.

The difficulty faced by the best observer is summarized by the Luenberger complexity
\begin{equation}
    Q(\widetilde A)
    =
    \frac{1}{1-\rho(\widetilde A)}
    \max\{1,\log \kappa_{\mathrm{diag}}(\widetilde A)\},
    \qquad
    Q_\star
    =
    \inf_L Q(A-LC).
    \label{eq:luenberger-complexity-background}
\end{equation}
This quantity depends on the spectral gap and diagonalization conditioning of the best closed-loop observer, rather than directly on the hidden-state dimension.

With $k$ spectral filters and finite memory order $n$, OSF uses the predictor
\begin{equation}
\begin{aligned}
    \hat y_t^{\mathrm{OSF}}
    =
    &\sum_{r=1}^{n} J_r u_{t-r}
    +
    \sum_{j=1}^{k}
    \sigma_j^{1/4} M_j
    U_{t,j}^{(u)}
    +
    \sum_{r=1}^{n} P_r y_{t-r}
    +
    \sum_{j=1}^{k}
    \sigma_j^{1/4} N_j
    U_{t,j}^{(y)} ,
\end{aligned}
\label{eq:osf-input-output-background}
\end{equation}
where
\begin{equation}
    U_{t,j}^{(u)}
    =
    \left\langle \phi_j,u_{t-1:t-L}\right\rangle,
    \qquad
    U_{t,j}^{(y)}
    =
    \left\langle \phi_j,y_{t-1:t-L}\right\rangle .
    \label{eq:osf-features-background}
\end{equation}
Here
$J_r,M_j\in\mathbb{R}^{d_y\times d_u}$ and
$P_r,N_j\in\mathbb{R}^{d_y\times d_y}$.
The parameters $(J,M,P,N)$ are learned over a bounded convex constraint set. Since the features are fixed, the prediction is linear in all trainable parameters, and squared loss remains convex. For nonlinear controlled dynamics,
\[
    x_{t+1}=f(x_t,u_t),
    \qquad
    y_t=g(x_t),
\]
OSF is used as an improper learner: it does not recover the latent state or fit a nonlinear model, but learns a predictor directly from filtered histories of inputs and observations, competing with the best finite-dimensional linear observer over the horizon. Under the boundedness and Lipschitz assumptions of \cite{dogariu2025universal}, choosing $k=\Theta(\log^2 T)$ yields an average prediction bound of the form
\begin{equation}
    \frac{1}{T}
    \sum_{t=1}^{T}
    \|y_t-\hat y_t^{\mathrm{OSF}}\|^2
    \le
    \widetilde O\!\left(
        \frac{
        Q_\star^2\log(Q_\star)\log^7 T
        }{\sqrt T}
    \right),
    \label{eq:osf-background-bound}
\end{equation}
after converting the bounded regret guarantee to squared loss. The important point for our purposes is that the rate is controlled by observer complexity $Q_\star$, not by the dimension of the finite-dimensional linear observer used in the comparison argument.

\subsection{Spectral-to-LDS distillation}

OSF learns a predictor written in terms of length-$L$ convolutions. Spectral-to-LDS distillation converts those convolutional features into recurrent coordinates.

For $\alpha_i\in[-1,1]$ and a signal stream $s_t$, define
\begin{equation}
    z_{t+1}^{(s,i)}
    =
    \alpha_i z_t^{(s,i)}
    +
    s_t .
    \label{eq:geometric-state-background}
\end{equation}
The implicit convolution of this recurrence is
\[
    (1,\alpha_i,\alpha_i^2,\ldots,\alpha_i^{L-1}) .
\]
Running $h$ such recurrences in parallel gives a dictionary of geometric filters. SpectraLDS chooses modes
$\alpha_1,\ldots,\alpha_h$ and a mixing matrix $F\in\mathbb{R}^{k\times h}$ such that
\begin{equation}
    \Phi_{1:k}
    \approx
    F\,\mu_L(\alpha_{1:h}),
    \label{eq:spectral-to-geometric-background}
\end{equation}
where $\Phi_{1:k} = [\phi_1, \dots, \phi_k]$ stacks the spectral filters and $\mu_L(\alpha_{1:h}) = [\mu_L(\alpha_1), \dots, \mu_L(\alpha_h)]$ stacks the geometric responses.

Therefore each spectral feature can be replaced by a linear combination of recurrent states:
\begin{equation}
    \left\langle \phi_j,s_{t-1:t-L}\right\rangle
    \approx
    \sum_{i=1}^{h}
    F(j,i)\,z_t^{(s,i)} .
    \label{eq:feature-distillation-background}
\end{equation}
Applying this replacement to both $s_t=u_t$ and $s_t=y_t$ in
\eqref{eq:osf-input-output-background} turns the learned OSF predictor into a recurrent LDS. The finite-memory terms
$\sum_{r=1}^{m}J_r u_{t-r}$ and $\sum_{r=1}^{m}P_r y_{t-r}$ are implemented by adding recurrent connections.

The two stages are complementary: OSF learns the predictor in a convex spectral feature space, and distillation realizes its temporal filters as recurrent coordinates. The resulting LDS's prediction error decomposes into the OSF learning error and the cost of replacing spectral-filter convolutions by recurrent geometric responses.

\section{Nonlinear-to-LDS Distillation}
\label{sec:unified}
\Cref{alg:nonlinear-to-lds} states the full pipeline for the controlled input-output setting. We fix
horizon $T$ and filter length $L=T$. 

\begin{algorithm}[H]
\caption{Nonlinear-to-LDS Distillation}
\label{alg:nonlinear-to-lds}
\begin{algorithmic}[1]
\REQUIRE Inputs $u_{1:T}$, observations $y_{1:T}$, filters $k$, AR order $n$, LDS modes $h$, filter length $L=T$
\STATE \textbf{Spectral learning.} Run OSF \cite{dogariu2025universal} with $k$ spectral filters
$\phi_1,\dots,\phi_k$ and AR order $n$ on the controlled trajectory $(u_{1:T},y_{1:T})$ to learn
the readout parameters $(J_r,M_j,P_r,N_j)$ and obtain $\hat y_t^{\mathrm{OSF}}$ as in
\eqref{eq:osf-input-output-background}.
\STATE \textbf{Distillation.} Apply SpectraLDS \cite{shah2025spectralds} to
$\Phi_{1:k}=[\phi_1\cdots\phi_k]^\top$ to obtain modes
$\alpha_1,\dots,\alpha_h\in[-1,1]$ and $F\in\mathbb{R}^{k\times h}$ such that
\[
    \Phi_{1:k} \approx F\,\mu_L(\alpha_{1:h}).
\]
\STATE \textbf{Recurrent realization.} Initialize
$z_1^{(u,i)}=0\in\mathbb{R}^{d_u}$ and
$z_1^{(y,i)}=0\in\mathbb{R}^{d_y}$ for $i=1,\dots,h$. At prediction time $t$, the states
$z_t^{(u,i)}$ and $z_t^{(y,i)}$ contain only the histories up to time $t-1$. Form
\[
    \widetilde U_{t,j}^{(u)}
    =
    \sum_{i=1}^h F(j,i) z_t^{(u,i)},
    \qquad
    \widetilde U_{t,j}^{(y)}
    =
    \sum_{i=1}^h F(j,i) z_t^{(y,i)} .
\]
\STATE \textbf{Output.} Predict
\[
\hat y_t^{\mathrm{LDS}}
=
\sum_{r=1}^n J_r u_{t-r}
+
\sum_{j=1}^k \sigma_j^{1/4} M_j \widetilde U_{t,j}^{(u)}
+
\sum_{r=1}^n P_r y_{t-r}
+
\sum_{j=1}^k \sigma_j^{1/4} N_j \widetilde U_{t,j}^{(y)} .
\]
\STATE \textbf{State update.} After time $t$ is processed, update
\[
    z_{t+1}^{(u,i)}
    =
    \alpha_i z_t^{(u,i)} + u_t,
    \qquad
    z_{t+1}^{(y,i)}
    =
    \alpha_i z_t^{(y,i)} + y_t .
\]
The finite-memory AR terms are implemented by shift registers over past inputs and observations.
The resulting predictor is an LDS with $O((h+m)(d_u+d_y))$ scalar state variables.
\end{algorithmic}
\end{algorithm}
 
\newtheorem{theorem}{Theorem}[section]
 
 \begin{theorem}[Nonlinear-to-LDS distillation]
\label{thm:nonlinear-to-lds}
Let $(f,g)$ be a nonlinear dynamical system s.t.:
\begin{enumerate}
    \item For all $t$, the inputs, states, and outputs are bounded:
    $\|u_t\|,\|x_t\|,\|y_t\|\le R$,
    \item The dynamics $f$ and observations $g$ are $1$-Lipschitz,
    \item The system has an observable discretized linear lifting
    (see Appendix~\ref{app:assumption}),
\end{enumerate}
which are the assumptions of \cite{dogariu2025universal}. Run
\Cref{alg:nonlinear-to-lds} with filter length $L=T$,
$k=\Theta(\log^2 T)$ filters, and $h$ SpectraLDS modes with $k \le h \le 4k$. Then, with probability at least $1-\delta$,
\[
\frac{1}{T}\sum_{t=1}^T
\|y_t-\hat y_t^{\mathrm{LDS}}\|^2
\le
\widetilde O\!\left(
\frac{Q_\star^2\log(Q_\star)\log^7T}{\sqrt T}
+
\lambda_{\max}(F)^2
\exp\!\left(-\frac{2k}{\log T}\right)
\right).
\]
where $\widetilde O(\cdot)$
suppresses logarithmic factors in $T$ as well as fixed
problem-dependent constants including $R$, $D$, $m$, and dimension factors. The distilled predictor uses $O((d_{in} + d_{out})(h+m))$ memory and per-step computation.
\end{theorem}
 
The bound has two terms. The OSF learning term $\widetilde O(Q_\star^2 \log Q_\star \log^7 T/\sqrt T)$ depends on the Luenberger complexity of the best linear observer. The distillation term  $\lambda_{\max}(F)^2 \exp\!\left(-\frac{2k}{\log T}\right) $
is the cost of replacing the learned convolutional predictor by a recurrent one. SpectraLDS observes empirically that $\lambda_{\textrm{max}}(F)$ is bounded and decreasing in $h$ \cite[Appendix~A.2]{shah2025spectralds}. The proof of \Cref{thm:nonlinear-to-lds} is given in \Cref{subsec:proof-main}.

\section{Experiments: linear systems and behavior cloning}
\label{sec:experiments-linear-bc}

The central theoretical contribution of this work is the first provable method for extracting a best-in-hindsight LDS representation of a nonlinear system via convex optimization. We showcase this contribution in two regimes. First, we study synthetic LDS recovery, where the target class is itself a linear dynamical system. Second, we study MuJoCo \cite{todorov2012mujoco}  behavior cloning, where the data are generated by nonlinear closed-loop expert policies and performance is measured by rollout reward. In these experiments, we consider the STU and OSF without the recurrent connections $J_r$ and $P_r$ (i.e., $n=0$).

\subsection{Learning linear dynamical systems}
\label{subsec:linear-lds}

We first consider prediction from a ground-truth LDS. Each task samples a system
\[
x_{t+1}=Ax_t+Bu_t,\qquad y_t=Cx_t\]
and trains a model to predict  $y_t$ from a length-$m$ history. We evaluate on two LDS families: \emph{symmetric} LDSs, whose transition matrices have real eigenvalues, and \emph{asymmetric} LDSs, whose non-normal transitions typically have complex eigenvalues and oscillatory responses. We compare the STU $\to$ SpectraLDS pipeline against directly training a Diagonal LDS,  training General LDS, and against training the more-general Linear FIR baselines. For the STU, we use $k = 48$ spectral filters. Each architecture has a $+y$ variant, which is observational and receives past outputs, and a $-y$ variant, which receives only inputs. Thus, the STU $+y$ variant is equivalent to OSF. For STU, all reported losses are computed after SpectraLDS conversion, so the curves evaluate the Diagonal LDS rather than the intermediate convolutional predictor.

We sweep learning rates over $\{10^{-1},10^{-2},10^{-3},10^{-4},10^{-5}\}$, select the value minimizing the geometric mean of final test MSE over the symmetric and asymmetric tasks, and rerun the selected configuration over five seeds. Full data-generation, initialization, learning-rate sweep results, and optimization details are given in Appendix~\ref{app:linear-lds-protocol}.
\paragraph{Long-memory systems.}
For the most challenging setting, we choose  $d_{\mathrm{in}}=d_{\mathrm{out}}=16$, $d_{\mathrm{state}}=64$, $m=512$, and spectral radius $\rho(A)=0.999$. In this regime the impulse response decays slowly, so accurate prediction requires retaining information far into the past. We compare parameter-matched STU, Diagonal LDS, and General LDS models at roughly $24{,}600$ parameters, together with a Linear FIR baseline, which requires more than ten times as many parameters. We also include STU trained with Online Newton Step (ONS), which we can employ as the prediction is linear in the STU trainable parameters; in contrast, the directly trained LDS baselines remain non-convex in their recurrent parameterizations. We test the $+y$ variant of each architecture.

\begin{figure}[H]
    \centering
    \includegraphics[width=0.95\linewidth]{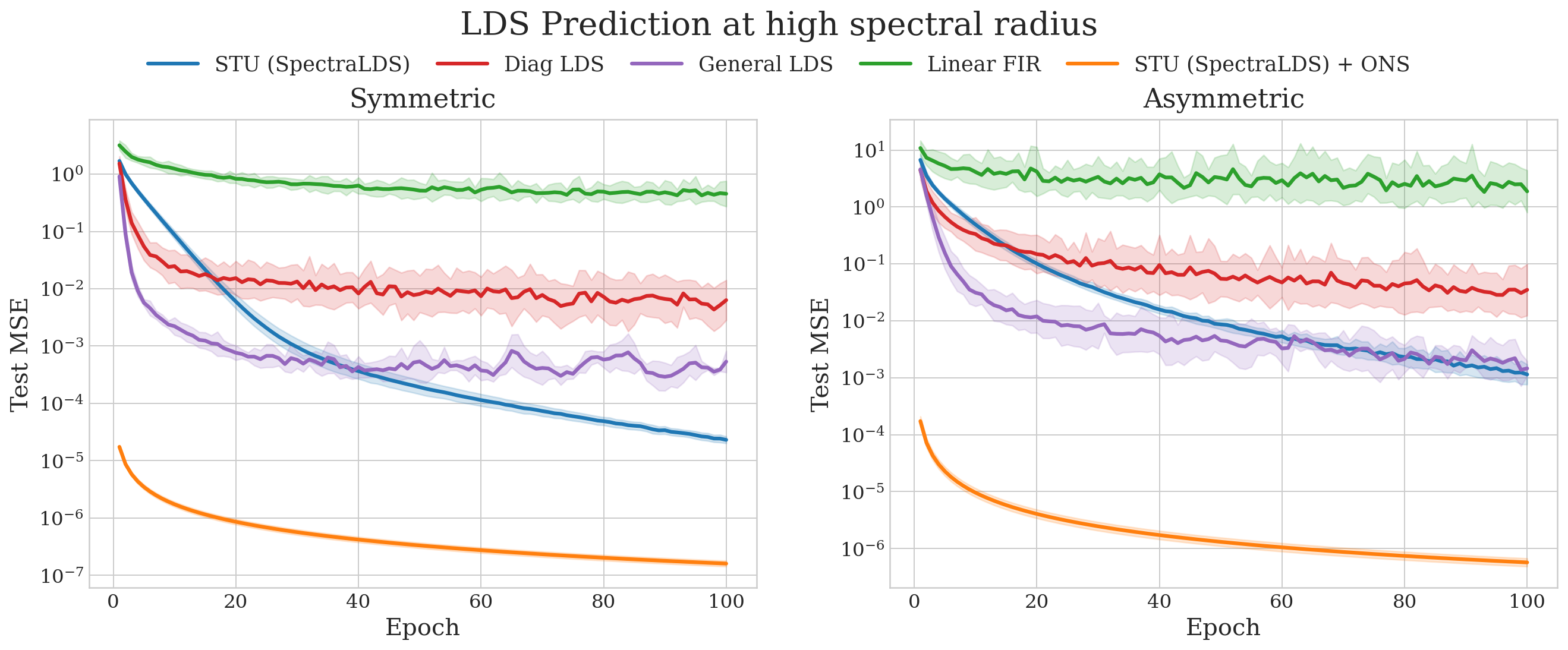}
    \caption{Test MSE versus epoch on the long-memory LDS task. Curves are geometric means over five seeds with geometric-standard-deviation bands, and the $y$-axis is log-scaled. STU $\to$ SpectraLDS is competitive under Adam and dominates when optimized with ONS, while Linear FIR fails despite having more than ten times as many parameters.}
    \label{fig:radius999}
\end{figure}

\Cref{tab:radius999} and \Cref{fig:radius999} show that STU $\to$ SpectraLDS outperforms each baseline under Adam, while STU $+$ ONS improves the error by another three orders of magnitude. The General LDS is more expressive than the Diagonal LDS, but it still does not close the gap to the spectral pipeline, suggesting that the main advantage is optimization rather than final model class. The wall-clock times in \Cref{tab:radius999} also show that the spectral model is substantially faster to train than the full General LDS, whose recurrent rollout dominates computation. Linear FIR requires $m \cdot (d_{in}+d_{out}) \cdot d_{out}$ parameters, leading to overparameterization that challenges optimization. In Appendix \ref{app:spectralds-fidelity}, we show that, in this setting, the STU and its distilled SpectraLDS representation have at most $1.5\%$ relative error.

\begin{table}[H]
\centering
\small
\caption{Linear LDS prediction in the long-memory regime ($d_{\mathrm{in}}{=}d_{\mathrm{out}}{=}16$, $d_{\mathrm{state}}{=}64$, $m{=}512$, $\rho(A){=}0.999$). Final test MSE over five seeds, parameter count, and wall-clock time for one 100-epoch run on an NVIDIA H100. STU rows report the SpectraLDS-distilled recurrent predictor.}
\label{tab:radius999}
\begin{tabular}{lrlll}
\toprule
Model & \# Params & Symmetric & Asymmetric & Time/run \\
\midrule
STU                                              & 24{,}576  & $2.3\!\times\!10^{-5} \pm 3.1\!\times\!10^{-6}$ & $1.3\!\times\!10^{-3} \pm 5.5\!\times\!10^{-4}$ &   5.4 s \\
Diag LDS (param-match)                           & 24{,}598  & $9.2\!\times\!10^{-3} \pm 8.9\!\times\!10^{-3}$ & $7.2\!\times\!10^{-2} \pm 1.0\!\times\!10^{-1}$ &  13.0 s \\
General LDS (param-match)                        & 24{,}705  & $6.0\!\times\!10^{-4} \pm 3.3\!\times\!10^{-4}$ & $1.5\!\times\!10^{-3} \pm 4.5\!\times\!10^{-4}$ & 403.6 s \\
Linear FIR                                       & 262{,}144 & $5.3\!\times\!10^{-1} \pm 3.4\!\times\!10^{-1}$ & $2.9\!\times\!10^{+0} \pm 2.9\!\times\!10^{+0}$ &   4.3 s \\
STU $+$ ONS ($\eta\!=\!1,\ \alpha\!=\!1$)        & 24{,}576  & $\mathbf{1.6\!\times\!10^{-7}} \pm 2.0\!\times\!10^{-8}$ & $\mathbf{5.7\!\times\!10^{-7}} \pm 8.7\!\times\!10^{-8}$ & 121.6 s \\
\bottomrule
\end{tabular}
\end{table}
\paragraph{Shorter-memory and larger-scale systems.}
We also test spectral radius $\rho(A)=0.95$, where transients decay much faster than in the long-memory experiment. We evaluate two scales:
\[
(d_{\mathrm{in}},d_{\mathrm{out}},d_{\mathrm{state}},m)
=
(4,4,8,256)
\quad\text{and}\quad
(64,64,64,512).
\]
At each scale we evaluate the $+y$ and $-y$ variants of STU $\to$ SpectraLDS, parameter-matched Diagonal LDS, Diagonal LDS with $h=2d_{\mathrm{state}}$, General LDS with $h=2d_{\mathrm{state}}$, and Linear FIR.

\begin{figure}[H]
    \centering
    \includegraphics[width=0.95\linewidth]{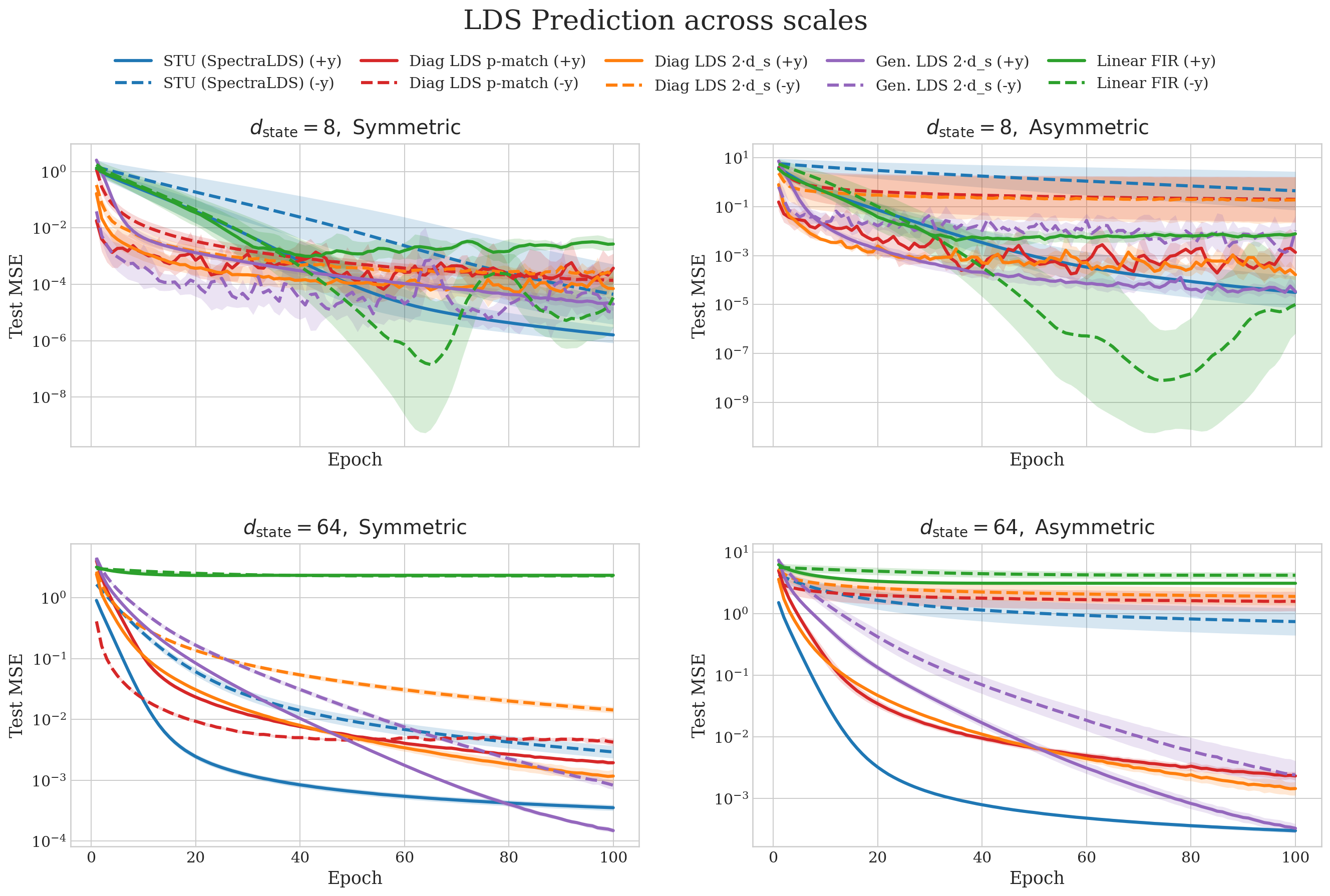}
    \caption{Test MSE versus epoch on linear LDS prediction at spectral radius $0.95$. The top row shows the small scale ($d_{\mathrm{in}}=d_{\mathrm{out}}=4$, $d_{\mathrm{state}}=8$, $m=256$), and the bottom row shows the large scale ($d_{\mathrm{in}}=d_{\mathrm{out}}=64$, $d_{\mathrm{state}}=64$, $m=512$). Solid curves are $+y$ variants and dashed curves are $-y$ variants. Curves are geometric means over five seeds with geometric-standard-deviation bands, and the $y$-axis is log-scaled.}
    \label{fig:cross-scale}
\end{figure}

For the large-scale experiment, we report the General LDS at $h=2d_{\mathrm{state}}$ rather than using a fully parameter-matched General LDS.  As shown in Table \ref{tab:radius999}, General LDS is substantially more expensive than the other methods, making rollouts of a parameter-matched General LDS model prohibitively expensive. We therefore use $h=2d_{\mathrm{state}}$ as the feasible General LDS baseline. For the Diagonal LDS, Figure \ref{fig:cross-scale} shows that the parameter-matched and $h=2d_{\mathrm{state}}$ variants have similar final test MSE. Full parameter counts are reported in Appendix~\ref{app:linear-lds-dims}.

\Cref{fig:cross-scale} shows that the qualitative pattern persists. At both scales, STU $\to$ SpectraLDS remains a strong predictor on both symmetric and asymmetric dynamics, although surpassed by Linear FIR on the smaller setting and General LDS at the larger setting. The diagonal LDS models, SpectraLDS and Diagonal LDS, without the $+y$ observation struggle on the asymmetric systems, where real diagonal recurrences cannot represent the complex modes of generic non-symmetric transitions. Linear FIR is competitive only in the smallest symmetric setting without the $+y$ observation, where its parameter count is not yet a bottleneck.

\subsection{MuJoCo behavior cloning}
\label{subsec:mujoco}

We next test the same comparison on nonlinear control data. For each MuJoCo \texttt{-v5} task \cite{todorov2012mujoco}, we roll out a deterministic Soft Actor-Critic (SAC) \cite{haarnoja2018softactorcriticoffpolicymaximum} expert trained by the Minari team \cite{minari} for $T=2{,}000{,}000$ steps and train a student to predict the expert action from a length-$m$ window of past observations and previous actions. The trained student is then deployed in the environment and evaluated by episode reward. Note that training uses teacher-forced next-action prediction on expert trajectories, while evaluation is closed-loop reward in the environment. MuJoCo results therefore serve as empirical evidence that the train-then-distill recipe survives this distribution shift, not as a direct test of our prediction-error bound, which concerns open-loop prediction.

All students share the same encoder, decoder, optimizer, batch size, and training horizon. The only component that changes is the sequence-mixing block:
\begin{enumerate}
    \item \textbf{STU $\to$ SpectraLDS}: trained as a tensor-dot STU approximation with $k=16$ spectral filters and deployed after conversion to a Diagonal LDS.
    \item \textbf{Diagonal LDS, hidden-dimension matched}: a directly trained Diagonal LDS with hidden dimension $h=d$.
    \item \textbf{Diagonal LDS, parameter-matched}: a directly trained Diagonal LDS whose hidden dimension is chosen to match the STU sequence-block parameter count.
\end{enumerate}
Every reward reported for STU is therefore the reward of the distilled recurrent LDS, not the pre-distillation convolutional model. All models are trained for $100$ epochs with Adam at learning rate $3\!\times\!10^{-4}$ and batch size $1024$, with five seeds per environment-model pair.

The students are intentionally compact: all three student architectures use fewer parameters than the SAC actor deployed by the teacher on every environment. Within the student models, STU and the parameter-matched Diagonal LDS have essentially identical parameter counts by construction, while the hidden-dimension-matched Diagonal LDS is roughly twenty to twenty-five percent larger. See Appendices~\ref{app:mujoco-arch} and~\ref{app:mujoco-params} for full details.

\begin{figure}[H]
    \centering
    \includegraphics[width=0.95\linewidth]{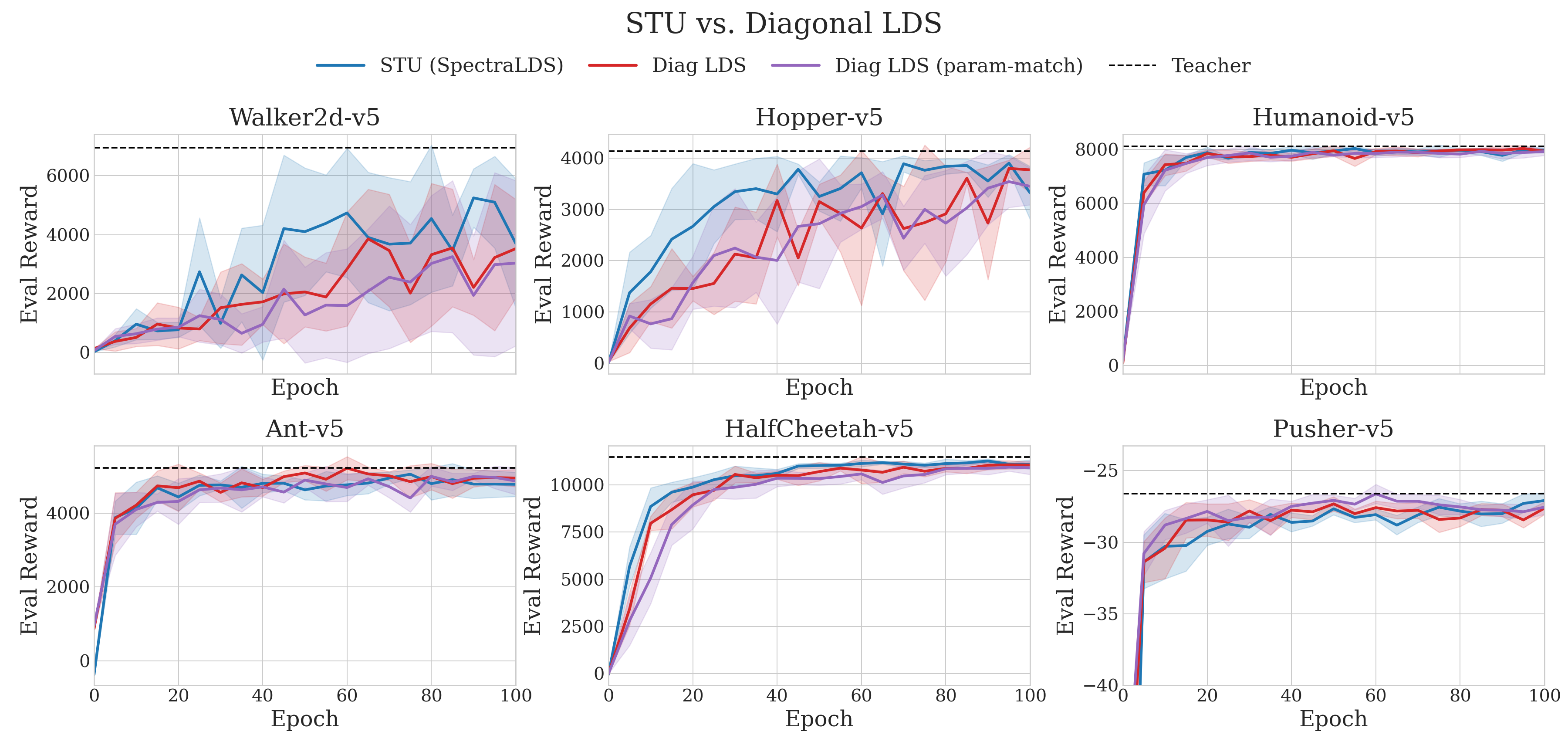}
    \caption{Evaluation reward during behavior cloning on six MuJoCo tasks. Curves show mean and standard deviation over five seeds. STU denotes the deployed SpectraLDS-distilled recurrent model.}
    \label{fig:mujoco-curves}
\end{figure}

\Cref{tab:mujoco-best} and \Cref{fig:mujoco-curves} show that STU $\to$ SpectraLDS achieves the highest mean reward on four of the six environments: Walker2d, Hopper, HalfCheetah, and Humanoid. On Ant and Pusher, it remains close to the best directly trained Diagonal LDS baseline. The largest separation occurs on Walker2d, where the spectral pipeline exceeds the parameter-matched Diagonal LDS by roughly $1{,}700$ reward. These results support the same conclusion as the LDS recovery experiments: training in the spectral parameterization and then distilling to an LDS is often more reliable than directly training the recurrent LDS, even when the final deployed model class is the same.

\begin{table}[H]
\centering
\small
\begin{tabular}{l c c c c}
\toprule
Environment & Teacher & STU $\to$ SpectraLDS & DiagLDS ($h\!=\!d$) & DiagLDS (param-match) \\
\midrule
Walker2d-v5    & 6956.4   & \textbf{6465.7} $\pm$ 329.8 & 5842.3 $\pm$ 555.0 & 4765.6 $\pm$ 2045.2 \\
Hopper-v5      & 4134.7   & \textbf{4037.6} $\pm$ 46.6  & 4006.8 $\pm$ 41.3   & 3927.5 $\pm$ 127.1  \\
Ant-v5         & 5224.1   & 5178.4 $\pm$ 99.8           & \textbf{5195.6} $\pm$ 167.7 & 5103.2 $\pm$ 272.3 \\
HalfCheetah-v5 & 11471.2  & \textbf{11198.4} $\pm$ 57.0 & 10972.3 $\pm$ 231.4 & 10838.1 $\pm$ 205.3 \\
Humanoid-v5    & 8110.3   & \textbf{8021.2} $\pm$ 70.4  & 7922.4 $\pm$ 88.6   & 7947.9 $\pm$ 89.5   \\
Pusher-v5      & $-26.6$  & $-27.4 \pm 0.5$             & $\mathbf{-27.1}$ $\pm$ 0.3 & $-27.3 \pm 0.6$ \\
\bottomrule
\end{tabular}
\caption{Best-checkpoint MuJoCo behavior-cloning reward, re-evaluated over $1{,}000$ episodes (mean $\pm$ std, five seeds). STU denotes the deployed SpectraLDS-distilled recurrent model.}
\label{tab:mujoco-best}
\end{table}

\section{Conclusion}

We showed that a convex spectral predictor can be distilled into a recurrent LDS whose prediction error decomposes into a distillation term that decays exponentially in the LDS state dimension and an OSF learning term controlled by the observer condition number $Q_\star$, independent of the latent nonlinear dimension. Empirically, the distilled LDS matches or outperforms directly trained LDS baselines at equal parameter budgets on both linear recovery and MuJoCo behavior cloning, with the advantage appearing to come from the spectral parameterization's convexity rather than from a richer model class.

Several directions remain open. On the theory side, we expect that a refined choice of sampling distribution over the SpectraLDS modes $\alpha_i$ would yield sharper guarantees on $\lambda_{\max}(F)$ in the overparameterized regime $h \gtrsim k$, where the current bound is dominated by the pseudoinverse norm. Additionally, the bound in Theorem~\ref{thm:nonlinear-to-lds} controls average one-step prediction error under teacher forcing. It does not cover closed-loop rollout error, where errors compound over the horizon --- the regime relevant to control and long-horizon simulation. Closing this gap is the most natural extension. Empirically, larger-scale experiments and integration with modern sequence models may establish nonlinear-to-LDS distillation as a building block for long-horizon forecasting and control.

\bibliographystyle{abbrv}
\bibliography{main}
 
\appendix

\section{Proof of Theorem~\ref{thm:nonlinear-to-lds}}
\label{subsec:proof-main}

The proof has two ingredients. First, the OSF guarantee gives a mean-squared
prediction bound for the learned spectral predictor. Second, replacing the OSF
filters by their SpectraLDS approximation perturbs the predictions by an amount
controlled by the filter approximation error and the norm of the learned readout.
Combining these two bounds gives the result.

Throughout, $\widetilde O(\cdot)$ suppresses logarithmic factors in $T$ and
fixed problem-dependent constants. We state the proof for scalar outputs for
readability; the vector-output case follows by using operator norms for the
readout matrices and contributes only the corresponding output-dimension factor.

\paragraph{Step 1: OSF mean-squared error.} 
By Theorem~3.1 of \cite{dogariu2025universal},
\begin{equation}
A_T
:=
\frac{1}{T}\sum_{t=1}^T
\|y_t-\hat y_t^{\mathrm{OSF}}\|^2
\le
\widetilde O\!\left(
\frac{Q_\star^2\log(Q_\star)\log^7T}{\sqrt T}
\right).
\label{eq:osf-msq}
\end{equation}

\paragraph{Step 2: SpectraLDS filter approximation.}
Run Algorithm~2 of \cite{shah2025spectralds} on $\Phi_{1:k}$ to obtain modes $\alpha_1,\dots,\alpha_h$ and a transformation matrix $F\in\mathbb R^{k\times h}$ (denoted $M_f$ in \cite{shah2025spectralds}).  We write
\[
\widetilde\phi_j = \sum_{i=1}^h F(j,i)\,\mu_L(\alpha_i), \qquad j=1,\dots,k.
\]
Define the rowwise approximation error
\begin{equation}
\eta_{k,h}(L) := \max_{j\le k} \|\phi_j - \widetilde\phi_j\|_1 .
\label{eq:eta-def}
\end{equation}
By Theorem~1 of \cite{shah2025spectralds}, on its high-probability event,
\begin{equation}
\eta_{k,h}(L)
\;\lesssim\;
h\,\lambda_{\max}(F)\,\exp\!\left(-\frac{k}{\log L}\right),
\label{eq:spectralds-filter-bound}
\end{equation}
where $\lambda_{\max}(F)$ is the largest singular value of the Moore--Penrose pseudoinverse of the $h\times k$ coefficient matrix produced by Algorithm~2 (called $M$ in \cite{shah2025spectralds}, with $F=M^+$).  As shown in \cite[Appendix~A.2]{shah2025spectralds}, the product $h\,\lambda_{\max}(F)$ is empirically bounded for the sampling distribution over $\alpha_i$ used by Algorithm~2.
\paragraph{Step 3: Filter error to prediction error.}
Let $\hat y_t^{\mathrm{dist}}$ be the predictor obtained by replacing each
$\phi_j$ in the OSF predictor with $\widetilde \phi_j$. As the finite-memory AR terms $\sum_r J_r u_{t-r}$ and $\sum_r P_r y_{t-r}$ in \eqref{eq:osf-input-output-background} are unchanged by the filter substitution and so do not contribute to the error, we have

\[
\hat y_t^{\mathrm{OSF}}-\hat y_t^{\mathrm{dist}}
=
\sum_{j=1}^k
\sigma_j^{1/4}M_j
\left\langle
\phi_j-\widetilde\phi_j,
u_{t-1:t-L}
\right\rangle
+
\sum_{j=1}^k
\sigma_j^{1/4}N_j
\left\langle
\phi_j-\widetilde\phi_j,
y_{t-1:t-L}
\right\rangle .
\]
Using $\|u_t\|,\|y_t\|\le R$ and
$\|\sigma_j^{1/4}M_j\|_{\mathrm{op}},
\|\sigma_j^{1/4}N_j\|_{\mathrm{op}}\le D$, we have
\[
\left\|
\left\langle
\phi_j-\widetilde\phi_j,
u_{t-1:t-L}
\right\rangle
\right\|
\le
R\|\phi_j-\widetilde\phi_j\|_1
\le
R\eta_{k,h}(L),
\]
and the same bound holds for the observation-history term. Therefore,
\[
\|\hat y_t^{\mathrm{OSF}}-\hat y_t^{\mathrm{dist}}\|
\le
2kDR\,\eta_{k,h}(L).
\]
Averaging over time gives the bound
\begin{equation}
\frac{1}{T}\sum_{t=1}^T
\|\hat y_t^{\mathrm{OSF}}-\hat y_t^{\mathrm{dist}}\|^2
\le
4k^2D^2R^2\,\eta_{k,h}(L)^2 .
\label{eq:dist-perturbation}
\end{equation}

Substituting \eqref{eq:spectralds-filter-bound} with $L=T$ yields
\begin{equation}
\frac{1}{T}\sum_{t=1}^T
\|\hat y_t^{\mathrm{OSF}}-\hat y_t^{\mathrm{dist}}\|^2
\le
\widetilde O\!\left(
k^2D^2R^2h^2\lambda_{\max}(F)^2
\exp\!\left(-\frac{2k}{\log T}\right)
\right).
\label{eq:dist-final-correct}
\end{equation}

The predictor $\hat y_t^{\mathrm{dist}}$ is realized exactly by an LDS: each
geometric response $\mu_L(\alpha_i)$ is implemented by a scalar recurrent mode,
and the AR terms are implemented by finite shift registers. We denote this exact recurrent realization by
$\hat y_t^{\mathrm{LDS}}$.

\paragraph{Step 4: Combine.}
For every $t$,
\[
y_t-\hat y_t^{\mathrm{LDS}}
=
(y_t-\hat y_t^{\mathrm{OSF}})
+
(\hat y_t^{\mathrm{OSF}}-\hat y_t^{\mathrm{dist}}).
\]
Using
\[
\|a+b\|^2\le 2\|a\|^2+2\|b\|^2,
\]
together with \eqref{eq:osf-msq} and \eqref{eq:dist-perturbation}, gives
\[
\frac{1}{T}\sum_{t=1}^T
\|y_t-\hat y_t^{\mathrm{LDS}}\|^2
\le
2A_T
+
8k^2D^2R^2\,\eta_{k,h}(L)^2 .
\]
Finally, substituting the SpectraLDS high-probability bound
\eqref{eq:spectralds-filter-bound} with $L=T$ gives the explicit estimate
\[
\frac{1}{T}\sum_{t=1}^T
\|y_t-\hat y_t^{\mathrm{LDS}}\|^2
\le
\widetilde O\!\left(
\frac{Q_\star^2\log(Q_\star)\log^7T}{\sqrt T}
+
k^2D^2R^2h^2\lambda_{\max}(F)^2
\exp\!\left(-\frac{2k}{\log T}\right)
\right).
\]

Since $k=\Theta(\log^2T)$, the factor $k^2$ is polylogarithmic in $T$ and is
absorbed by $\widetilde O(\cdot)$. The constants $D$ and $R$, together with $h = O(k)$, are also absorbed by
$\widetilde O(\cdot)$. Thus the same bound can be written in the cleaner form
\[
\frac{1}{T}\sum_{t=1}^T
\|y_t-\hat y_t^{\mathrm{LDS}}\|^2
\le
\widetilde O\!\left(
\frac{Q_\star^2\log(Q_\star)\log^7T}{\sqrt T}
+
\lambda_{\max}(F)^2
\exp\!\left(-\frac{2k}{\log T}\right)
\right).
\]
 This proves the theorem. \hfill$\square$

\section{Assumption: Observable Discretized Linear Lifting}
\label{app:assumption}

We recall the discretized lifting assumption used in \cite{dogariu2025universal}.
For simplicity, we state the autonomous version and consider:
\[
    x_{t+1}=f(x_t), \qquad y_t=g(x_t),
\]

Fix a discretization scale $\varepsilon>0$, and let
$S=\{s^{(1)},\dots,s^{(N)}\}$ be an $\varepsilon$-net of the radius-$R$ ball in
state space. Thus every bounded state $x$ has a representative
$\pi(x)\in S$ with
\[
    \|x-\pi(x)\|\le \varepsilon,
    \qquad
    N=|S|\le \left(\frac{2R}{\varepsilon}\right)^{d_X}.
\]
Define a finite-state approximation to the nonlinear dynamics by
\[
    T(s)=\pi(f(s)).
\]
Equivalently, define a transition matrix $A'\in\mathbb R^{N\times N}$ by
\[
    A'_{ij}
    =
    \mathbf 1\{T(s^{(j)})=s^{(i)}\}.
\]
The lifted state $z_t\in\mathbb R^N$ is the one-hot vector representing the
current grid point. Its evolution is linear:
\[
    z_{t+1}=A'z_t.
\]
Define the observation matrix $C'\in\mathbb R^{d_Y\times N}$ by setting its
$j$th column to the observation at the corresponding grid point:
\[
    C' e_j = g(s^{(j)}).
\]
The discretized linear lifting is therefore the finite-dimensional LDS
\[
    z_{t+1}=A'z_t,
    \qquad
    y'_t=C'z_t.
\]

By Lemma~5.5 of \cite{dogariu2025universal}, if $f$ and $g$ are
$1$-Lipschitz and the trajectory is bounded, then this lifted LDS approximates
the nonlinear output sequence over horizon $T$:
\[
    \sum_{t\le T}\|y'_t-y_t\|
    \le
    \frac{T^2}{2}\varepsilon .
\]
Thus, choosing $\varepsilon$ sufficiently small produces a high-dimensional LDS
whose outputs track the nonlinear trajectory. The dimension
$N\le (2R/\varepsilon)^{d_X}$ can be very large, but OSF is an improper learner:
its runtime and parameter count do not depend on explicitly constructing or
learning this lifted state.

The additional assumption in Theorem~\ref{thm:nonlinear-to-lds} is that the pair
$(A',C')$ is observable. That is, the observability matrix
\[
    \mathcal O(A',C')
    =
    \begin{bmatrix}
        C'\\
        C'A'\\
        C'(A')^2\\
        \vdots\\
        C'(A')^{N-1}
    \end{bmatrix}
    \in \mathbb R^{Nd_Y\times N}
\]
has full column rank:
\[
    \operatorname{rank}\mathcal O(A',C')=N.
\]
Equivalently, no nonzero lifted state direction is invisible to all future
linear observations:
\[
    C'(A')^\ell v=0
    \quad
    \text{for all } \ell=0,\dots,N-1
    \quad
    \Longrightarrow
    \quad
    v=0.
\]

This observability condition is used only in the comparison argument. It ensures
that a Luenberger observer can be formed for the lifted LDS and that the
observer complexity $Q_\star$ appearing in the OSF bound is finite.

\section{Linear LDS data-generation and training protocol}
\label{app:linear-lds-protocol}

This appendix documents the data-generation procedure, predictor
initializations, and training
protocol used in the linear-LDS experiments of
Section~\ref{subsec:linear-lds}.

\subsection{Data generation}
\label{app:linear-lds-data}

Each ground-truth LDS $(A, B, C)$ is sampled once per (seed, task) pair.
\begin{itemize}
    \item \textbf{Symmetric} $A$: draw $Q \in \mathbb{R}^{d_\mathrm{state} \times d_\mathrm{state}}$ from the QR
    decomposition of a Gaussian matrix and eigenvalues
    $\lambda_i \stackrel{\text{i.i.d.}}{\sim} \mathrm{Uniform}(-\rho, \rho)$;
    set $A = Q\,\mathrm{diag}(\lambda)\,Q^\top$.
    \item \textbf{Asymmetric} $A$: draw $\tilde A_{ij} \stackrel{\text{i.i.d.}}{\sim} \mathcal{N}(0, 1/d_\mathrm{state})$ and
    rescale to the target spectral radius via
    $A = \rho\,\tilde A / \max_i |\lambda_i(\tilde A)|$. The eigenvalues are
    typically complex, producing oscillatory dynamics.
    \item $B \in \mathbb{R}^{d_\mathrm{state} \times d_\mathrm{in}}$ and
    $C \in \mathbb{R}^{d_\mathrm{out} \times d_\mathrm{state}}$ have entries
    $\mathcal{N}(0, 1/d_\mathrm{in})$ and $\mathcal{N}(0, 1/d_\mathrm{state})$, respectively.
\end{itemize}
Inputs are sampled i.i.d.\ as $u_t \sim \mathcal{N}(0, I_{d_\mathrm{in}})$, and the
state evolves from $x_0 = 0$ via $x_{t+1} = A x_t + B u_t$ with output
$y_t = C x_t$. We generate $T_{\mathrm{train}} = 10{,}000$ training time
steps and $T_{\mathrm{test}} = 2{,}000$ test time steps; the test inputs
are drawn with a different RNG seed from the same distribution, so train
and test share the same dynamics but no shared input samples.

The predictor sees a length-$m$ sliding window of past observations and
predicts the current output. From a trajectory of length $T$ this yields
$N = T - m$ examples; the input at index $i$ is
$(u_{i+1}, \dots, u_{i+m})$ together with, in the $+y$ variant, the
corresponding past outputs $(y_{i}, \dots, y_{i+m-1})$, and the target is
$y_{i+m}$. So the training set contains
$N_{\mathrm{tr}} = T_{\mathrm{train}}-m$ samples and the held-out test
set contains $T_{\mathrm{test}} - m$ samples.

\subsection{Predictor initializations}
\label{app:linear-lds-init}

All predictors are strictly linear; the only nonlinearity in any of the
parameterizations is the $\tanh$ on Diagonal LDS eigenvalues. We use
default PyTorch dtypes (float32 throughout training) and the random
initialization given here, applied with the same seed used for the
ground-truth LDS so that any seed-to-seed variance reflects a single
jointly resampled configuration of data and model.
\begin{itemize}
    \item \textbf{STU } (with $k$ spectral filters): the
    spectral filters $\phi \in \mathbb{R}^{m \times k}$ are computed
    deterministically from the $m \times m$ Hankel matrix and held fixed. 
    The mixer $M \in \mathbb{R}^{k \times d_\mathrm{in} \times d_\mathrm{out}}$
    is initialized i.i.d.\ $\mathcal{N}(0, 1/(k\cdot d_\mathrm{in}))$. For the experiments with $m = 512$, we choose SpectraLDS $h = 160$, and otherwise we choose $h = 120$.
    \item \textbf{Diagonal LDS} with hidden dimension $h$: the raw
    eigenvalue parameter is $\tilde\lambda \sim \mathrm{Uniform}(-2,2)^h$,
    and the trained eigenvalues are
    $\lambda = \tanh(\tilde\lambda) \cdot 0.999$, which keeps every
    $\lambda_i \in (-0.999, 0.999)$ for the entire training run and so
    guarantees stability without needing projection or clipping. The
    input matrix $B \in \mathbb{R}^{h \times d_\mathrm{in}}$ has entries
    $\mathcal{N}(0, 1/d_\mathrm{in})$ and the output matrix
    $C \in \mathbb{R}^{d_\mathrm{out} \times h}$ has entries $\mathcal{N}(0, 1/h)$. We train the Diagonal LDS model convolutionally for maximal efficiency.
    \item \textbf{General LDS} with hidden dimension $h$:
    $A \in \mathbb{R}^{h \times h}$ is drawn from PyTorch's
    \texttt{nn.init.orthogonal\_} and then rescaled in-place by
    $\rho_{\mathrm{init}} / \rho(A)$ with $\rho_{\mathrm{init}} = 0.95$
    so that $\rho(A) = 0.95$ at initialization. After this, $A$ is an
    unconstrained \texttt{nn.Parameter}; we do not project, clip, or
    otherwise normalize $A$ during training, and rely only on the
    per-step gradient-norm clip described below to bound updates. The
    input and readout matrices use the same scheme as the Diagonal LDS.
    \item \textbf{Linear FIR}: the per-tap kernel
    $W \in \mathbb{R}^{m \times d_\mathrm{out} \times d_\mathrm{in}^{(\pm y)}}$ has entries
    $\mathcal{N}(0, 1/(m\, d_\mathrm{in}^{(\pm y)}))$, where
    $d_\mathrm{in}^{(\pm y)} = d_\mathrm{in} + d_\mathrm{out}$ in the $+y$ variant and
    $d_\mathrm{in}$ in the $-y$ variant. The factor of $m$ in the variance is
    necessary to keep the predicted output's variance $O(1)$ at
    initialization; without it, the $m$-tap sum makes early-epoch
    predictions blow up by roughly $\sqrt{m}$ and Adam's first updates
    are catastrophically large.
\end{itemize}

\subsection{Optimization and hyperparameter sweep}
\label{app:linear-lds-opt}

\paragraph{First-order baseline (Adam).}
All Adam-trained models are run for 100 epochs at batch size 256 on the
small-scale and high-spectral-radius experiments and batch size 64 at the
large scale (where the longer windows and larger hidden dimensions make
the larger batch run out of memory). After each backward pass we apply
\texttt{torch.nn.utils.clip\_grad\_norm\_(model.parameters(), 1.0)}. Adam's other hyperparameters are
PyTorch defaults ($\beta_1 = 0.9$, $\beta_2 = 0.999$, $\varepsilon = 10^{-8}$,
weight decay $0$). For each (model, $\pm y$) we sweep the learning rate
at seed $0$ over the grid
$\eta_{\mathrm{Adam}} \in \{10^{-1}, 10^{-2}, 10^{-3}, 10^{-4}, 10^{-5}\}$,
select the value minimizing the geometric mean of final-epoch test MSE
across the symmetric and asymmetric tasks, and rerun at seeds
$\{0, 1, 2, 3, 4\}$ with that LR. 

\paragraph{Second-order learner for STU (ONS, used in the
high-spectral-radius experiment).}
The STU is linear in its parameters. Collecting the spectral features into the flattened vector
$g
\in \mathbb{R}^{k\cdot d_\mathrm{in}^{(\pm y)}}$ where $g = \text{vec}(g_{full})$ for $g_{full}[n,i] = \sum_t x_{\mathrm{rev}}[t,i]\, \phi[t,n]$, the predictor reduces to $\hat y = M^\top g$ for a flattened mixer
$M$. Thus, for MSE loss, Online Newton Step
\citep{hazan-newton} attains $O(\log T)$ regret. We use the standard ONS
update
\[
M \leftarrow M - \eta\, A_t^{-1}\, g_t\, (M^\top g_t - y_t)^\top,
\]
with the inverse Hessian estimate maintained by Sherman--Morrison from
$A_t = \alpha I + \sum_{s \le t} g_s g_s^\top$. We chose $\eta = 1$ and fixed
$\alpha  = 1$ after observing instability with our earlier $\alpha = 1\times10^{-3}$ default. Thus the final choice is $(\eta, \alpha) = (1, 1)$.

\subsection{Per-experiment dimensions}
\label{app:linear-lds-dims}

The two cross-scale settings and the high-spectral-radius setting share
the protocol above; only the dimensions differ.

\begin{table}[h]
\centering
\small
\begin{tabular}{l c c c c c c}
\toprule
Experiment & $d_\mathrm{in}$ & $d_\mathrm{out}$ & $d_\mathrm{state}$ & $m$ & $\rho$ & batch \\
\midrule
Cross-scale, small  & 4  & 4  & 8  & 256 & $0.95$  & 256 \\
Cross-scale, large  & 64 & 64 & 64 & 512 & $0.95$  & 64  \\
High spectral radius & 16 & 16 & 64 & 512 & $0.999$ & 256 \\
\bottomrule
\end{tabular}
\caption{Dimensions used in each experiment. The number of spectral
filters $k$ is fixed at $48$ throughout; for the parameter-matched
Diagonal and General LDS variants, the hidden dimension $h$ is chosen so
that the predictor matches STU's parameter count
($k (d_\mathrm{in}^{(\pm y)})\, d_\mathrm{out}$) at each scale.}
\label{tab:linear-lds-dims}
\end{table}

In all settings the training set has
$N_{\mathrm{tr}} = T_{\mathrm{train}} - m$ samples and the test set has
$T_{\mathrm{test}} - m$ samples; with $T_{\mathrm{train}} = 10{,}000$
and $T_{\mathrm{test}} = 2{,}000$, the longest-window cases ($m = 512$:
cross-scale large and high spectral radius) leave $9{,}488$ training
windows and $1{,}488$ test windows.
 
\subsection{Hyperparameter sweep}
\label{app:linear-lds-sweep}

In this section, we include the results of the learning rate hyperparameter sweep for the learning linear dynamical systems experiments in Section \ref{sec:experiments-linear-bc}.

\begin{table}[h]                                                                                         
  \centering                                                                                             
  \caption{Learning-rate sweep at spectral radius $\rho=0.999$                                             
  ($d_{\mathrm{in}}{=}d_{\mathrm{out}}{=}16$, $d_{\mathrm{state}}{=}64$, $m{=}512$, 100 epochs, seed 0).
  Final test MSE on symmetric and asymmetric tasks. For STU+ONS, we fixed $\eta = 1$ and tested $\alpha\in\{1, 1^{-3}\}$, choosing $\alpha = 1$.}                                                         
  \label{tab:rho0999_lr_sweep}                                                                             
  \begin{tabular}{l l c c}                                                                                 
  \toprule                                                                                                 
  Method & LR / $\eta$ & Sym.\ test MSE & Asym.\ test MSE \\
  \midrule                                                                                                 
  \multirow{5}{*}{STU + $y$ (Adam)}
    & $10^{-5}$ & $1.48\!\times\!10^{0}$  & $9.23\!\times\!10^{0}$ \\                                      
    & $10^{-4}$ & $6.89\!\times\!10^{-2}$ & $4.25\!\times\!10^{-1}$ \\                                     
    & $10^{-3}$ & $\mathbf{2.44\!\times\!10^{-5}}$ & $\mathbf{9.61\!\times\!10^{-4}}$ \\                   
    & $10^{-2}$ & $1.08\!\times\!10^{-3}$ & $9.57\!\times\!10^{-3}$ \\                                     
    & $10^{-1}$ & $9.64\!\times\!10^{-1}$ & $1.51\!\times\!10^{0}$ \\                                      
  \midrule                                                                                                 
  \multirow{5}{*}{Diag LDS + $y$ (Adam)}                                                                   
    & $10^{-5}$ & $4.59\!\times\!10^{-1}$ & $1.99\!\times\!10^{0}$ \\                                      
    & $10^{-4}$ & $\mathbf{2.86\!\times\!10^{-3}}$ & $1.17\!\times\!10^{-1}$ \\                            
    & $10^{-3}$ & $7.71\!\times\!10^{-3}$ & $\mathbf{2.10\!\times\!10^{-2}}$ \\                            
    & $10^{-2}$ & $8.23\!\times\!10^{-3}$ & $8.47\!\times\!10^{-2}$ \\                                     
    & $10^{-1}$ & $1.37\!\times\!10^{0}$  & $8.35\!\times\!10^{0}$ \\                                      
  \midrule                                                                                                 
  \multirow{5}{*}{General LDS + $y$ (Adam)}                                                                
    & $10^{-5}$ & $6.35\!\times\!10^{-1}$ & $3.27\!\times\!10^{0}$ \\          
    & $10^{-4}$ & $8.97\!\times\!10^{-4}$ & $1.10\!\times\!10^{-2}$ \\
    & $10^{-3}$ & $\mathbf{3.73\!\times\!10^{-4}}$ & $\mathbf{2.12\!\times\!10^{-3}}$ \\                   
    & $10^{-2}$ & $6.45\!\times\!10^{-3}$ & $6.06\!\times\!10^{-2}$ \\          
    & $10^{-1}$ & NaN & NaN \\                       
  \midrule                       \multirow{5}{*}{Linear FIR + $y$ (Adam)}
    & $10^{-5}$ & $1.96\!\times\!10^{0}$  & $5.04\!\times\!10^{0}$ \\                                      
    & $10^{-4}$ & $6.70\!\times\!10^{-1}$ & $1.19\!\times\!10^{0}$ \\                                      
    & $10^{-3}$ & $\mathbf{2.59\!\times\!10^{-1}}$ & $\mathbf{1.19\!\times\!10^{0}}$ \\                    
    & $10^{-2}$ & $2.64\!\times\!10^{1}$  & $1.90\!\times\!10^{2}$ \\          
    & $10^{-1}$ & $4.15\!\times\!10^{3}$  & $2.25\!\times\!10^{4}$ \\                             
  \bottomrule                  
  \end{tabular}   
  \end{table}

\begin{table}[h]
\centering
\small
\caption{Learning-rate sweep at the
small scale ($d_{\mathrm{in}}{=}d_{\mathrm{out}}{=}4$, $d_{\mathrm{state}}{=}8$,
$m{=}256$) and the large scale ($d_{\mathrm{in}}{=}d_{\mathrm{out}}{=}d_{\mathrm{state}}{=}64$,
$m{=}512$). We report final test MSE after $100$-epochs of training at seed $0$ with Adam. Bold marks the best LR per model and scale.}
\label{tab:cross_scale_lr_sweep}
\begin{tabular}{l c c c c c}
\toprule
& & \multicolumn{2}{c}{$d_{state}{=}8$ } & \multicolumn{2}{c}{$d_state{=}64$} \\
\cmidrule(lr){3-4}\cmidrule(lr){5-6}
Method & LR & Sym & Asym & Sym & Asym \\
\midrule
\multirow{5}{*}{STU + $y$ (Adam)}
  & $10^{-1}$ & $2.61\!\times\!10^{-4}$ & $5.79\!\times\!10^{-2}$ & $2.72\!\times\!10^{0}$  & $5.05\!\times\!10^{0}$ \\
  & $10^{-2}$ & $1.69\!\times\!10^{-5}$ & $5.72\!\times\!10^{-7}$ & $2.11\!\times\!10^{-2}$ & $4.30\!\times\!10^{-2}$ \\
  & $10^{-3}$ & $\mathbf{1.08\!\times\!10^{-6}}$ & $\mathbf{8.16\!\times\!10^{-6}}$ & $\mathbf{3.33\!\times\!10^{-4}}$ & $\mathbf{2.84\!\times\!10^{-4}}$ \\
  & $10^{-4}$ & $1.14\!\times\!10^{-1}$ & $1.83\!\times\!10^{-1}$ & $1.15\!\times\!10^{-2}$ & $2.29\!\times\!10^{-2}$ \\
  & $10^{-5}$ & $9.07\!\times\!10^{-1}$ & $3.07\!\times\!10^{0}$  & $8.95\!\times\!10^{-1}$ & $1.66\!\times\!10^{0}$ \\
\midrule
\multirow{5}{*}{Diag LDS p-match + $y$ (Adam)}
  & $10^{-1}$ & $1.39\!\times\!10^{-2}$ & $3.07\!\times\!10^{-3}$ & $1.16\!\times\!10^{2}$  & $2.71\!\times\!10^{2}$ \\
  & $10^{-2}$ & $\mathbf{5.39\!\times\!10^{-4}}$ & $\mathbf{1.90\!\times\!10^{-4}}$ & $6.92\!\times\!10^{-3}$ & $3.36\!\times\!10^{-2}$ \\
  & $10^{-3}$ & $7.85\!\times\!10^{-5}$ & $2.03\!\times\!10^{-3}$ & $4.08\!\times\!10^{-3}$ & $1.09\!\times\!10^{-2}$ \\
  & $10^{-4}$ & $6.73\!\times\!10^{-4}$ & $1.19\!\times\!10^{-3}$ & $\mathbf{1.83\!\times\!10^{-3}}$ & $\mathbf{2.42\!\times\!10^{-3}}$ \\
  & $10^{-5}$ & $3.35\!\times\!10^{-1}$ & $1.33\!\times\!10^{0}$  & $6.89\!\times\!10^{-2}$ & $1.56\!\times\!10^{-1}$ \\
\midrule
\multirow{5}{*}{Diag LDS $2{\cdot}d_{\mathrm{state}}$ + $y$ (Adam)}
  & $10^{-1}$ & $1.76\!\times\!10^{-3}$ & $1.11\!\times\!10^{-2}$ & $4.60\!\times\!10^{0}$  & $9.13\!\times\!10^{0}$ \\
  & $10^{-2}$ & $\mathbf{4.32\!\times\!10^{-4}}$ & $\mathbf{2.21\!\times\!10^{-4}}$ & $3.90\!\times\!10^{-2}$ & $8.17\!\times\!10^{-2}$ \\
  & $10^{-3}$ & $1.12\!\times\!10^{-3}$ & $2.07\!\times\!10^{-3}$ & $\mathbf{8.64\!\times\!10^{-4}}$ & $\mathbf{1.21\!\times\!10^{-3}}$ \\
  & $10^{-4}$ & $3.27\!\times\!10^{-2}$ & $1.25\!\times\!10^{-1}$ & $4.01\!\times\!10^{-3}$ & $1.05\!\times\!10^{-2}$ \\
  & $10^{-5}$ & $1.99\!\times\!10^{0}$  & $4.83\!\times\!10^{0}$  & $8.27\!\times\!10^{-1}$ & $1.43\!\times\!10^{0}$ \\
\midrule
\multirow{5}{*}{Gen.\ LDS $2{\cdot}d_{\mathrm{state}}$ + $y$ (Adam)}
  & $10^{-1}$ & NaN & NaN & NaN & NaN \\
  & $10^{-2}$ & $3.30\!\times\!10^{-6}$ & $7.94\!\times\!10^{-5}$ & $3.47\!\times\!10^{-2}$ & $1.52\!\times\!10^{-1}$ \\
  & $10^{-3}$ & $\mathbf{7.24\!\times\!10^{-6}}$ & $\mathbf{2.69\!\times\!10^{-5}}$ & $1.40\!\times\!10^{-3}$ & $2.53\!\times\!10^{-3}$ \\
  & $10^{-4}$ & $4.94\!\times\!10^{-3}$ & $5.64\!\times\!10^{-3}$ & $\mathbf{1.56\!\times\!10^{-4}}$ & $\mathbf{4.15\!\times\!10^{-4}}$ \\
  & $10^{-5}$ & $2.97\!\times\!10^{0}$  & $6.93\!\times\!10^{0}$  & $3.33\!\times\!10^{-1}$ & $8.05\!\times\!10^{-1}$ \\
\midrule
\multirow{5}{*}{Linear FIR + $y$ (Adam)}
  & $10^{-1}$ & $3.38\!\times\!10^{1}$  & $7.00\!\times\!10^{1}$  & $1.16\!\times\!10^{5}$  & $3.38\!\times\!10^{5}$ \\
  & $10^{-2}$ & $3.33\!\times\!10^{-1}$ & $8.90\!\times\!10^{-1}$ & $1.16\!\times\!10^{3}$  & $3.37\!\times\!10^{3}$ \\
  & $10^{-3}$ & $\mathbf{1.18\!\times\!10^{-3}}$ & $\mathbf{8.02\!\times\!10^{-3}}$ & $1.37\!\times\!10^{1}$  & $3.73\!\times\!10^{1}$ \\
  & $10^{-4}$ & $1.11\!\times\!10^{-1}$ & $2.60\!\times\!10^{-1}$ & $2.39\!\times\!10^{0}$  & $3.67\!\times\!10^{0}$ \\
  & $10^{-5}$ & $9.13\!\times\!10^{-1}$ & $2.36\!\times\!10^{0}$  & $\mathbf{2.29\!\times\!10^{0}}$  & $\mathbf{3.34\!\times\!10^{0}}$ \\
\bottomrule
\end{tabular}
\end{table}

\section{MuJoCo behavior-cloning data and training protocol}
\label{app:mujoco-arch}

This appendix documents the teacher source, dataset construction, shared architecture, sequence-mixing-block details, training hyperparameters, and evaluation protocol used in the MuJoCo experiments of Section~\ref{subsec:mujoco}.

\subsection{Environments}
\label{app:mujoco-envs}

We evaluate the train-then-distill pipeline on six MuJoCo \texttt{-v5} tasks spanning a range of locomotion and manipulation difficulties. Table~\ref{tab:mujoco-envs-app} lists each environment alongside its observation and action dimensions, along with the history length $m$, embedding dimension $d$, and number of spectral filters $k$ used by the student. For SpectraLDS distillation, we use $h = 80$.

\begin{table}[H]
    \centering
    \small
    \begin{tabular}{l|cccccc}
    \hline
    Environment & Description & $d_o$ & $d_a$ & $m$ & $d$ & $k$ \\
    \hline
    \texttt{HalfCheetah-v5} & 2D quadruped with the goal of running           & $17$  & $6$  & $8$   & $64$  & $16$ \\
    \texttt{Hopper-v5}      & 2D monoped with the goal of hopping             & $11$  & $3$  & $128$ & $128$ & $16$ \\
    \texttt{Walker2d-v5}    & 2D biped with the goal of walking               & $17$  & $6$  & $128$ & $128$ & $16$ \\
    \texttt{Ant-v5}         & 3D quadruped with the goal of running           & $105$ & $8$  & $8$   & $64$  & $16$ \\
    \texttt{Humanoid-v5}    & 3D biped with the goal of walking               & $348$ & $17$ & $128$ & $128$ & $16$ \\
    \texttt{Pusher-v5}      & 3D arm with the goal of pushing an object       & $23$  & $7$  & $128$ & $128$ & $16$ \\
    \hline
    \end{tabular}
    \vspace{5px}
    \caption{The six MuJoCo environments used in our behavior-cloning experiments, listed with their observation dimension $d_o$, action dimension $d_a$, history length $m$, and the used embedding dimension $d$. Each student receives the concatenation $[o_t, a_{t-1}]$ at each time step, so the input dimension to the encoder is $d_o + d_a$.  Environment descriptions are adapted from the Minari documentation \cite{minari} of the MuJoCo environments \cite{todorov2012mujoco}.}
    \label{tab:mujoco-envs-app}
\end{table}

\subsection{Teachers and trajectory data}
\label{app:mujoco-data}

For each of the six environments we distill a single Soft Actor-Critic \cite{haarnoja2018softactorcriticoffpolicymaximum} teacher policy trained by the Minari team \cite{minari}. 

Each teacher is rolled out for $T = 2{,}000{,}000$ time steps in its environment using a fixed seed and the deterministic action mode. The resulting trajectory $(o_t, a_t)_{t=0}^{T-1}$ is the only data the student ever sees: no rollouts of the student itself appear in the training distribution. From this trajectory we construct a sliding-window supervised dataset
\[
\mathcal{D} \;=\; \big\{\, \big( (o_{t-m+1:t},\, a_{t-m:t-1}),\; a_t \big) \,:\, t = m, \dots, T-1 \,\big\},
\]
in which the student observes a length-$m$ window of past observations and previous actions at each time step and is trained, with mean-squared error against the teacher's action $a_t$, to predict the next action. We do not subsample or shuffle the trajectory, so the entire $T - m$ samples are used as training data for every student.

The ``test" set in this experiment is the live environment itself: trained students are deployed deterministically and evaluated on $N_{\mathrm{eval}} = 100$ episodes per checkpoint. We report best-checkpoint reward, defined as the mean episode reward of the checkpoint that achieves the highest mean reward over its $1{,}000$ deployment episodes. We save checkpoints every 5 epochs. Each episode runs to the environment's default time limit or natural termination under the standard \texttt{-v5} reward function.

\subsection{Shared encoder, head, and action squash}
\label{app:mujoco-shared}

To isolate the effect of the sequence-mixing block from the surrounding architecture, all three students share an identical encoder and decoder architecture with embedding dimension $d$ that depends on the environment, (see Table~\ref{tab:mujoco-envs-app}).

\begin{itemize}
    \item \textbf{Encoder} ($\mathbb{R}^{d_o + d_a} \to \mathbb{R}^{d}$): a two-layer GELU MLP applied independently at each time step. Both hidden layers have width $d$, no normalization layers, and GELU is applied between the two linear layers. The encoder concatenates the current observation with the previous action before applying the MLP, so each time step in the sliding window contributes a single $d$-dimensional embedding to the sequence-mixing block's input.
    \item \textbf{Sequence-mixing block} ($\mathbb{R}^{m \times d} \to \mathbb{R}^{d}$): the only component that varies across the three students. It consumes the length-$m$ sequence of encoder embeddings and produces a single $d$-dimensional summary. The three variants are detailed in Appendix~\ref{app:mujoco-mixing-blocks}.
    \item \textbf{Decoder} ($\mathbb{R}^{d} \to \mathbb{R}^{d_a}$): a three-layer GELU MLP with first layer $d$, second layer $d/2$, and output $d_a$, with  GELU between layers and a final \texttt{tanh} normalization and scaling to ensure the student's predictions live in the valid action box.
\end{itemize}

The only architectural variation across the three students lies in the sequence-mixing block.

\subsection{Sequence-mixing-block variants}
\label{app:mujoco-mixing-blocks}

The three students differ only in their sequence-mixing block.

\paragraph{STU (the train-then-distill variant).}
At training time, the block is a tensor-dot Spectral Transform Unit \cite{liu2024flashstu} with $k = 16$ spectral filters.  We use the tensor-dot variant for parameter efficiency at $d = 128$; Appendix~\ref{app:stu-canonical-vs-tensordot} compares it against the STU on the linear-LDS task. At evaluation time, the trained STU is converted to a Diagonal LDS via SpectraLDS, and the converted Diagonal LDS is what we deploy.  To allow for easier reuse of the published SpectraLDS conversions \cite{shah2025spectralds}, we use the spectral filters computed with $m = 8192$ and $k = 48$ and at $80$-dimensions, truncate, and consider the relevant subset of $k = 16$ filters. The STU runs in \texttt{float64} precision due to historical reasons, while other models are in the native \texttt{float32}. Based on later experimentation, we do not expect this contributes to any differences.

\paragraph{Diagonal LDS, $C$-matched ($h = d$).}
A directly-trained recurrence $x_{t+1} = \mathrm{diag}(\lambda)\, x_t + B u_t,\ y_t = C x_t$, with $\lambda = \tanh(\tilde\lambda)\,(1 - 10^{-3})$ to keep eigenvalues strictly inside the unit disk. 

\paragraph{Diagonal LDS, parameter-matched.}
Identical to the $C$-matched variant in form, but with $h$ chosen so that $h(1 + 2d) = 2kd + d^2$, giving exactly the same parameter count as STU's tensor-dot block. This is the parameter-matched baseline for the train-then-distill comparison.

\subsection{Per-student parameter counts}
\label{app:mujoco-params}

Table~\ref{tab:mujoco-params-app} reports the total trainable parameters of each student alongside the SAC teacher's parameter count. All three students are smaller than the SAC actor used at inference time on every environment, and roughly an order of magnitude smaller than the full SAC checkpoint (which carries two critic networks and their target copies in addition to the actor). Within the students, STU and the parameter-matched Diagonal LDS have essentially identical parameter counts by construction, while the $C$-matched Diagonal LDS is approximately twenty to twenty-five percent larger because the $h = d$ choice inflates its sequence-mixing block.

\begin{table}[H]
\centering
\small
\begin{tabular}{l c c c c c}
\toprule
Environment & STU & DiagLDS ($h\!=\!d$) & DiagLDS (param-match) & SAC actor & SAC full \\
\midrule
\texttt{Walker2d-v5}    & 62{,}150  & 77{,}638  & 62{,}218  & 73{,}484  & 362{,}256 \\
\texttt{Hopper-v5}      & 60{,}803  & 76{,}291  & 60{,}871  & 70{,}406  & 349{,}962 \\
\texttt{Ant-v5}         & 22{,}568  & 26{,}216  & 22{,}604  & 97{,}040  & 477{,}972 \\
\texttt{HalfCheetah-v5} & 16{,}742  & 20{,}390  & 16{,}778  & 73{,}484  & 362{,}256 \\
\texttt{Humanoid-v5}    & 106{,}641 & 122{,}129 & 106{,}709 & 163{,}874 & 802{,}854 \\
\texttt{Pusher-v5}      & 63{,}111  & 78{,}599  & 63{,}179  & 75{,}534  & 371{,}474 \\
\bottomrule
\end{tabular}
\caption{Total trainable parameters per student model and the corresponding SAC teacher across the six MuJoCo environments. The "SAC actor" column gives the parameter count of the policy network used at inference time, while "SAC full" adds the critic and target-critic networks carried in the SAC checkpoint and so reflects the size of the full SAC artifact. The SpectraLDS and parameter-matched LDS students use fewer parameters than the SAC actor on every environment, and within the students the $C$-matched Diagonal LDS is uniformly larger than the STU pipeline by roughly twenty to twenty-five percent.}
\label{tab:mujoco-params-app}
\end{table}

\subsection{Optimization and training protocol}
\label{app:mujoco-opt}

All three students are trained for $100$ epochs on the $T - m$-sample dataset using Adam at a fixed learning rate of $3 \times 10^{-4}$ and batch size $1{,}024$. We use PyTorch's default Adam hyperparameters otherwise ($\beta_1 = 0.9$, $\beta_2 = 0.999$, $\varepsilon = 10^{-8}$, weight decay $0$). We do not sweep the learning rate in MuJoCo; we use $3 \times 10^{-4}$ for every (environment, student) pair, and verified on a small set of pilot runs that all three students train stably at this rate. We run five seeds $\{0, 1, 2, 3, 4\}$ per environment-student pair.

Checkpoints are saved every five training epochs and at the end of training. Each checkpoint is evaluated on $100$ episodes, and the reward we report in Table~\ref{tab:mujoco-best} is from taking the best checkpoint and re-evaluating it on $1{,}000$ episodes.

\section{SpectraLDS distillation fidelity}
\label{app:spectralds-fidelity}

A key step in our evaluation pipeline is converting the trained STU predictor into a SpectraLDS (diagonal-LDS) predictor.  The filter fit is approximate, so we verify here that this approximation does not meaningfully change the test MSE.

\begin{figure}[t]
    \centering
    \includegraphics[width=\linewidth]{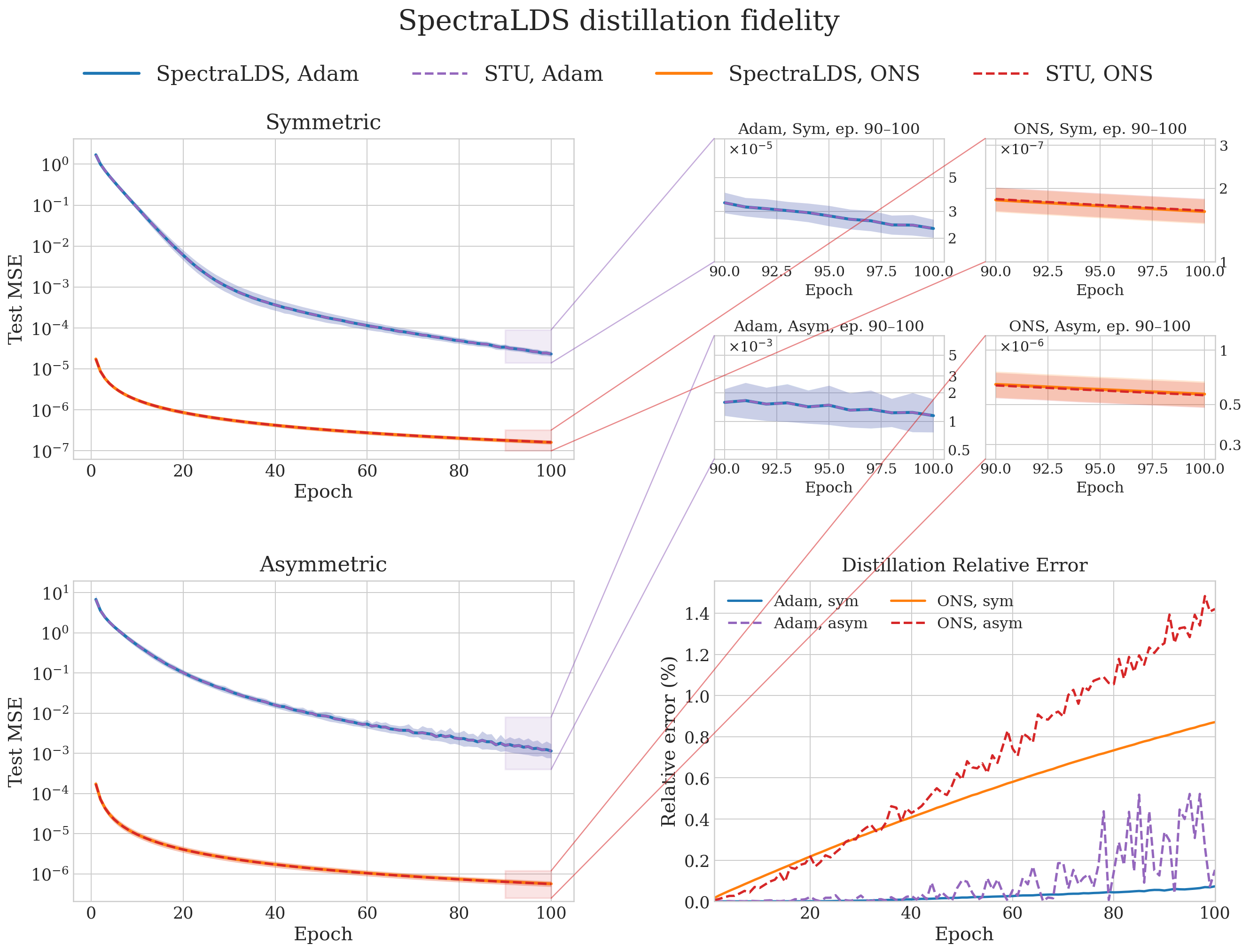}
    \caption{Test MSE versus epoch for the STU predictor (dashed)
    and the SpectraLDS-distilled predictor (solid), at
    $\rho = 0.999$. Each curve is the geometric mean over five seeds with
    geometric-standard-deviation bands; the $y$-axis is log-scaled.
    The solid and dashed curves overlap almost exactly in every panel,
    confirming that SpectraLDS distillation introduces negligible error.}
    \label{fig:spectralds-fidelity}
\end{figure}

Figure~\ref{fig:spectralds-fidelity} overlays the STU test MSE (from
the original five-seed runs) and the SpectraLDS-distilled test MSE (from
independent five-seed runs on the same seeds), for both the Adam and ONS
optimizers on the high-spectral-radius setting ($\rho = 0.999$, $d = 16$,
$d_{\mathrm{state}} = 64$, $m = 512$).  Each curve is the geometric mean
over five seeds with geometric-standard-deviation bands; the solid line
is the distilled model and the dashed line is the raw STU.

The two curves are visually indistinguishable in every panel. Quantitatively, the maximum relative error between the raw and distilled
test MSEs, taken over all epochs and all seeds, is below $1.5\%$ in all conditions. 

\section{STU vs.\ tensor-dot STU}
\label{app:stu-canonical-vs-tensordot}

Both STU variants begin from the same intermediate object. Given an
input window $x \in \mathbb{R}^{m \times d_{\mathrm{in}}}$ and the fixed
spectral filters $\phi \in \mathbb{R}^{m \times k}$, the
spectrally-filtered representation of the window is the matrix
\[
Z \;=\; \phi^\top x \;\in\; \mathbb{R}^{K \times d_{\mathrm{in}}},
\qquad
Z_{k, i} \;=\; \big\langle \phi_{:, k},\; x_{:, i} \big\rangle.
\]
$Z$ has one row per spectral filter and one column per input channel; all
of the time-window information that the predictor will use lives in this
$k \times d_{\mathrm{in}}$ matrix. Both STU variants then linearly map
$Z$ to a $d_{\mathrm{out}}$-dimensional prediction $y$. 

\paragraph{STU.}
The STU uses an unconstrained linear map from
$\mathbb{R}^{k \times d_{\mathrm{in}}}$ to $\mathbb{R}^{d_{\mathrm{out}}}$,
parameterized by a tensor
$M \in \mathbb{R}^{k \times d_{\mathrm{in}} \times d_{\mathrm{out}}}$:
\[
y \;=\; \sum_{k, i} M_{k, i, :}\; Z_{k, i}.
\]
Every element of $M$ is independently learned; the map has
$k\, d_{\mathrm{in}}\, d_{\mathrm{out}}$ free parameters.

\paragraph{Tensor-dot STU.}
The tensor-dot variant replaces the single mixer tensor
$M \in \mathbb{R}^{k \times d_{\mathrm{in}} \times d_{\mathrm{out}}}$
with two matrices,
\[
M_{\mathrm{in}} \in \mathbb{R}^{k \times d_{\mathrm{in}}},
\qquad
M_{\mathrm{out}} \in \mathbb{R}^{d_{\mathrm{in}} \times d_{\mathrm{out}}},
\]
constraining the mixer to factorize as
$M_{k, i, o} = M_{\mathrm{in}, k, i}\cdot M_{\mathrm{out}, i, o}$. Two
benefits follow.

First, the parameter count drops from $k\, d_{\mathrm{in}}\, d_{\mathrm{out}}$
to $k\, d_{\mathrm{in}} + d_{\mathrm{in}}\, d_{\mathrm{out}}$ — at the MuJoCo
dimensions $d_{\mathrm{in}} = d_{\mathrm{out}} = 128$ and $k = 16$ this is a
$7.5\times$ reduction.

Second, the factorization permits a cheaper convolutional implementation.
In the STU, predicting $y$ involves $k \cdot d_{\mathrm{in}}$
distinct filter convolutions. The tensor-dot factorization lets us instead
combine $M_{\mathrm{in}}$ with the spectral filters into
$d_{\mathrm{in}}$ per-input-channel filters
$f_{:, i} = \sum_k M_{\mathrm{in}, k, i}\, \phi_{:, k}$, then perform only
$d_{\mathrm{in}}$ time-convolutions (one per input channel) before mixing
through $M_{\mathrm{out}}$. The number of convolutions thus drops
from $k \cdot d_{\mathrm{in}}$ to $d_{\mathrm{in}}$.

However, the tensor-dot STU does have reduced representation capacity, which we analyze in the learning an LDS setting.

\begin{figure}[t]
    \centering
    \includegraphics[width=\linewidth]{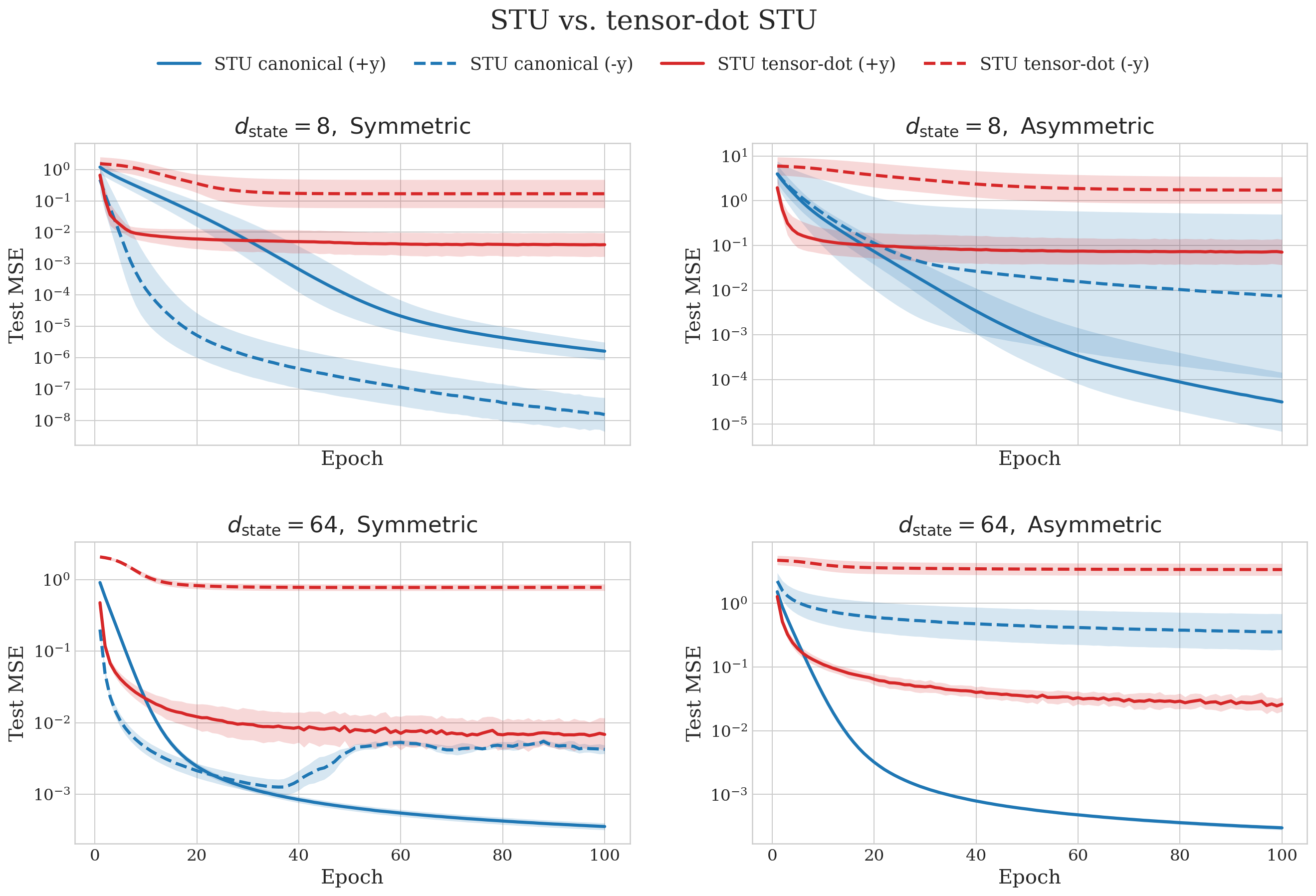}
    \caption{Test MSE versus epoch for STU (blue) and
    tensor-dot STU (red), both with $k = 48$, and each in $+y$ (solid) and $-y$ (dashed) variants,
    on the linear-LDS benchmark of
    Section~\ref{subsec:linear-lds}. Top row: small scale ($d=4$,
    $d_{\mathrm{state}}=8$, $m=256$). Bottom row: large scale ($d=64$,
    $d_{\mathrm{state}}=64$, $m=512$). Curves are geometric means over five
    seeds with geometric-standard-deviation bands; the y-axis is
    log-scaled.}
    \label{fig:stu-canonical-vs-tensordot}
\end{figure}

\paragraph{Comparison.}
Figure~\ref{fig:stu-canonical-vs-tensordot} reports test MSE versus
epoch for both STU variants in $+y$ and $-y$ form, at both the small
($d_{\mathrm{state}} = 8$, $m = 256$) and large ($d_{\mathrm{state}} = 64$,
$m = 512$) scales. Each curve is the geometric mean over five seeds with
geometric-standard-deviation bands, on a log-scaled vertical axis. The STU outperforms the tensor-dot approximation STU in each setting as expected, however, the tensor-dot STU has approximately $k\times$ fewer parameters.

\end{document}